\documentclass[sigconf]{acmart}
\AtBeginDocument{%
  }

\copyrightyear{2025}
\acmYear{2025}
\setcopyright{cc}
\setcctype{by}
\acmConference[KDD '25]{Proceedings of the 31st ACM SIGKDD Conference on
Knowledge Discovery and Data Mining V.2}{August 3--7, 2025}{Toronto, ON,
Canada}
\acmBooktitle{Proceedings of the 31st ACM SIGKDD Conference on Knowledge
Discovery and Data Mining V.2 (KDD '25), August 3--7, 2025, Toronto, ON,
Canada}
\acmDOI{10.1145/3711896.3737078}
\acmISBN{979-8-4007-1454-2/2025/08}

\author{Zhiyuan Zhao}
\email{leozhao1997@gatech.edu}
\affiliation{%
  \institution{Georgia Institute of Technology}
  \city{Atlanta}
  \state{GA}
  \country{USA}
}

\author{Haoxin Liu}
\email{hliu763@gatech.edu}
\affiliation{%
  \institution{Georgia Institute of Technology}
  \city{Atlanta}
  \state{GA}
  \country{USA}
}

\author{Alexander Rodríguez}
\email{alrodri@umich.edu}
\affiliation{%
  \institution{University of Michigan}
  \city{Ann Arbor}
  \state{MI}
  \country{USA}
}

\author{B. Aditya Prakash}
\email{badityap@cc.gatech.edu}
\affiliation{%
  \institution{Georgia Institute of Technology}
  \city{Atlanta}
  \state{GA}
  \country{USA}
}

\usepackage{amssymb}
\usepackage{algorithm}
\usepackage{algpseudocode}
\usepackage{xspace}
\usepackage{bm}
\usepackage{amsfonts}
\usepackage{amsthm}
\usepackage{mathtools}
\usepackage{multirow}
\usepackage{bbding}
\usepackage{pifont}
\usepackage{wasysym}
\usepackage{amsbsy}
\usepackage{url}
\usepackage{subcaption}

\usepackage{balance}

\newcommand{\problem}[0]{{\textsc{PeTS}}\xspace}
\newcommand{\solution}[0]{{\textsc{FPS}}\xspace}

\begin{document}

\title{Performative Time-Series Forecasting}



\begin{abstract}
Time-series forecasting is a critical challenge in various domains and has witnessed substantial progress in recent years. Many real-life scenarios, such as public health, economics, and social applications, involve feedback loops where predictive models can trigger actions that influence the outcome they aim to predict, subsequently altering the target variable's distribution. This phenomenon, known as \textit{performativity}, introduces the potential for 'self-negating' or 'self-fulfilling' predictions. Despite extensive studies on performativity in classification problems across domains, this phenomenon remains largely unexplored in the context of time-series forecasting from a machine-learning perspective.

In this paper, we formalize \textit{Performative Time-Series Forecasting} (\problem), addressing the challenge for predictions when performativity-induced distribution shifts are possible. We propose a novel solution to \problem, \textit{Feature Performative-Shifting} (\solution), which leverages the concept of delayed response to anticipate distribution shifts and subsequently predicts targets. 
We provide theoretical insights suggesting that \solution potentially leads to reduced generalization error. 
Extensive experiment results demonstrate that \solution consistently outperforms conventional time-series forecasting and concept drift methods, highlighting its efficacy in handling performativity-induced challenges.
\end{abstract}

\ccsdesc[500]{Computing methodologies~Neural networks}

\keywords{Time-Series Forecasting, Performativity, Distribution Shift}


\maketitle

\section{Introduction}
\label{sec:intro}



Time-series forecasting is a fundamental task in time-series analysis that finds applications in various domains such as economics, urban computing, and epidemiology~\cite{zhu2002statstream, zheng2014urban, rodriguez2022data, mathis2024evaluation}. These applications involve predicting future trends or events based on historical time-series data. For example, economists use forecasts to make financial and marketing plans, while sociologists use them to allocate resources and formulate policies for traffic or disease control.

The advent of deep learning methods has revolutionized time-series forecasting~\cite{lai2018modeling, torres2021deep, salinas2020deepar, Yuqietal-2022-PatchTST, zhou2021informer}. Despite existing models have achieved state-of-the-art performance by capturing long-term dependencies using well-designed attentions,
these models are developed without considering feedback loops between input features and targets. These models neglect the fact that predictions can influence the outcomes they aim to predict, a phenomenon referred to as \textit{performativity}~\cite{perdomo2020performative}. While performativity has been extensively studied in literature such as strategic classification and reinforcement learning~\cite{bartlett1992learning, lu2018learning, bhati2022performative}, it remains largely unexplored in time-series forecasting problems.

\begin{figure}[t]
    \centering
    \vspace{0.1in}
    \includegraphics[width=0.40\textwidth]{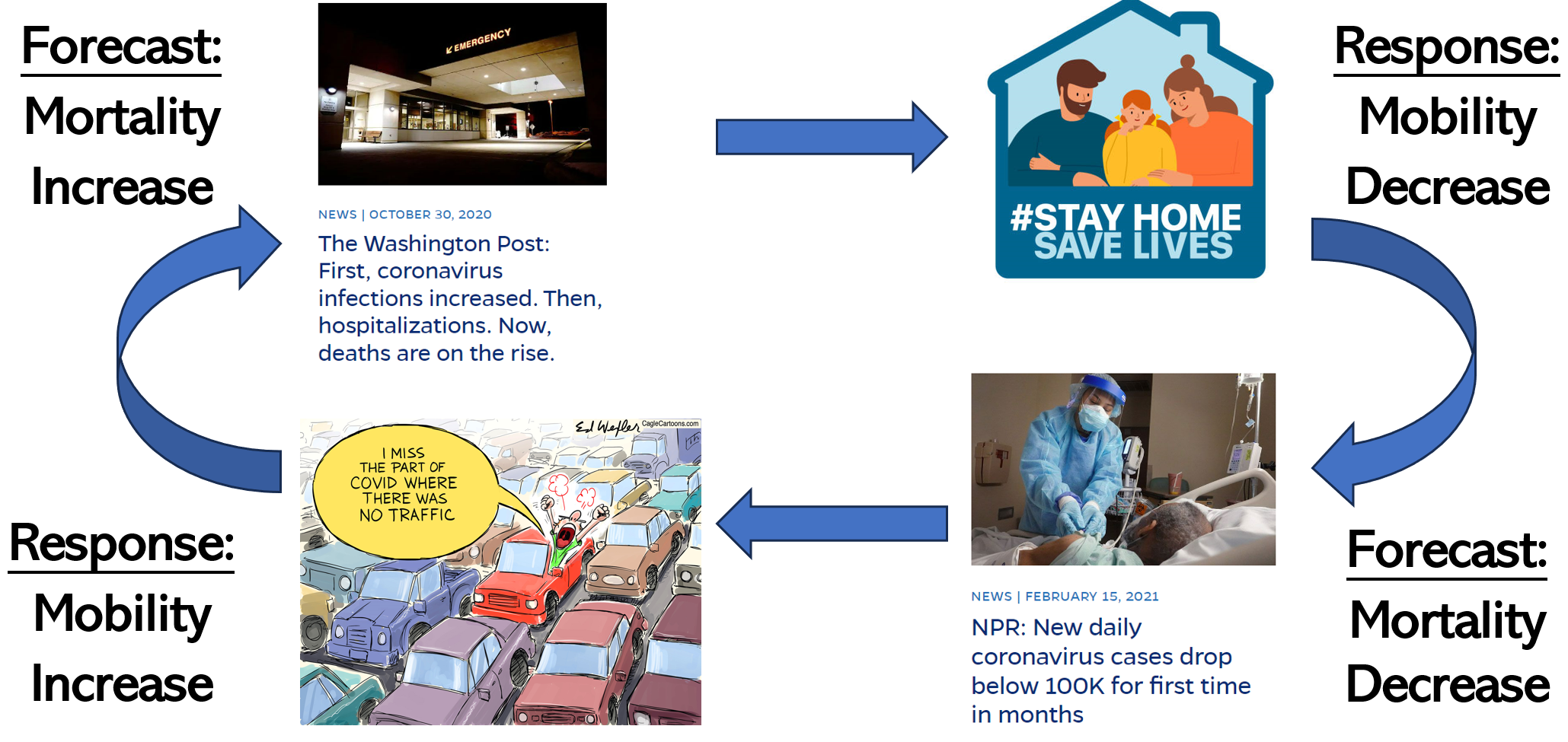}
    \caption{Example of performative feedback loop for COVID-19 mortality and mobility.}
    \label{fig:diag_intro}
\end{figure}

Performativity shows as `self-negating' or `self-fulfilling' predictions in real-world scenarios. For instance, traffic predictions can influence traffic patterns, and disease forecasts can affect behavior interventions, impacting future predictions. Figure \ref{fig:diag_intro} illustrates a performativity feedback loop between COVID-19 mortality predictions and mobility features. In this feedback loop, forecasting during a severe pandemic leads to more stringent interventions, resulting in reduced mobility and subsequently reduced mortality, and vice versa. Realizing the prevalence of performativity prompts studying its effects on time-series forecasting. Despite recent theoretical work of performativity in supervised learning for classification problems~\cite{perdomo2020performative, mendler2020stochastic, mandal2022performative, bhati2022performative}, performativity in time-series forecasting remains understudied both theoretically and practically. This paper formally defines the \underline{Pe}rformative \underline{T}ime-\underline{S}eries (\problem) forecasting problem, which aims to make robust predictions in the presence of performativity-induced distribution shifts. Our main contributions are:
\begin{enumerate} 
    \item \textbf{Novel Problem:} We introduce the novel performative time-series forecasting (\problem) problem, which focuses on providing robust forecasts under the influence of performativity~\cite{perdomo2020performative}. This refers to changes in the distribution of the target variable caused by the predictions themselves due to feedback loops.
    \item \textbf{New Methodology:} We propose a natural solution to \problem, \underline{F}eature \underline{P}erformative-\underline{S}hifting (\solution), which anticipates and incorporates responses resulting from performativity into the forecasting process. We provide theoretical insights into how using delayed responses contributes to better forecasting performance. Importantly, \solution is a model-agnostic framework that can be integrated with existing time-series forecasting models.
    \item \textbf{Extensive Experiments:} We evaluate \solution on various time-series models and datasets for both standard and real-time forecasting settings. Extensive evaluation results show benefits of \solution for solving \problem, as well as outperformance than existing concept drift methods, in forecasting accuracy and trend-change capturing ability.
\end{enumerate}


\section{Related Work}
\label{sec:related_work}


\par\noindent\textbf{Time-Series Forecasting.} Classical time series models such as ARIMA~\cite{hyndman2018forecasting} focus on univariate time series which predicts each time series independently. Other statistical methods treat time-series forecasting as standard regression problems with time-varying parameters, solving with conventional regression techniques, such as kernel regression, support vector regression, and Gaussian processes~\cite{nadaraya1964estimating, smola2004tutorial, williams1995gaussian}. 

Recent works in deep learning have achieved notable achievements in time-series forecasting, such as LSTNet, N-BEATS, and S4~\cite{lai2018modeling, Oreshkin2020:N-BEATS, gu2022efficiently}. Other deep learning methods have built upon the successes of self-attention mechanisms~\cite{vaswani2017attention} with transformer-based architectures, such as Informer, Autoformer, PatchTST, and more~\cite{zhou2021informer, wu2021autoformer, Yuqietal-2022-PatchTST, kamarthi2022camul, liu2024lstprompt}. These transformer-based models exhibit accurate forecasting accuracy through capturing long-range dependencies.

\par\noindent\textbf{Concept Drift.} Learning under non-stationary distributions, where the conditional distribution from exogenous covariates to the target changes with time, known as \textit{concept drift}, has attracted attention from the learning theory community~\cite{kuh1990learning, bartlett1992learning}. 
Concept shift methods~\cite{arjovsky2019invariant,ahuja2021invariance,VREx,pezeshki2021gradient,sagawa2019distributionally} assume instances sampled from diverse environments and propose to identify and utilize invariant predictors across environments. In time series data, concept shift represents a common yet challenging issue, given the temporal changes of mapping relationships and the challenges in acquiring adequate environment labels under the conventional time-series forecasting setups~\cite{liu2024time1}. Other recent studies have explored multimodal time series analysis, which enhances numerical forecasting by incorporating additional contextual information~\cite{liu2024time}. Such analyses introduce new reasoning capabilities that hold promise for improved causal inference and mitigating concept drift in time series~\cite{liu2024picture, liu2025can, Evaluating}.

\par\noindent\textbf{Performative Prediction.} Performativity severs a particular case of concept drifts that states the phenomenon that the predictive model influences the distribution of future covariates through responses taken based on the model's predictions. This phenomenon has been well-studied in domains including social sciences and economics~\cite{mackenzie2007economists, healy2015performativity}. In contrast to considering arbitrary shifts in general concept drifts, performative prediction problems, such as strategic classification~\cite{perdomo2020performative, mendler2020stochastic, brown2022performative}, elaborates concept drifts with specific orientations of distribution shifts and learning objectives.

\section{Preliminaries}
\label{sec:preliminaries}

Performative prediction involves making predictions where the predictive model $\theta$ affects the underlying data distribution $\mathcal{D}(\theta)$. Prior statistical works focus on a specific case problem of performative prediction called strategic classification~\cite{hardt2016strategic, ghalme2021strategic, perdomo2020performative, mendler2020stochastic, mandal2022performative}. In strategic classification, individuals strategically modify their input features before undergoing classification. The key challenge lies in anticipating the strategic responses to the classifier, drawn from the performativity-induced distribution $\mathcal{D}(\theta)$. This problem can be framed as an optimization problem:  
\begin{gather}
    \theta^* = \arg\min_{\theta} \mathop{\mathbb{E}}_{\mathcal{Z}\sim \mathcal{D}(\theta)} \mathcal{L} (\mathcal{Z};\theta)
\label{eqn:perf}
\end{gather}

Thus, the fundamental difficulty in strategic classification lies in capturing the distributional shift $\mathcal{D}(\cdot)$ induced by $\theta$. Existing works~\cite{perdomo2020performative, bhati2022performative} address this challenge by modeling strategic behavior as an optimization problem: $\pmb{x}_{br} = \arg\max_{\pmb{x}^\prime} u(\pmb{x}^\prime, \theta) - c(\pmb{x}^\prime, \pmb{x})$, where $\pmb{x}$ is the original feature, and $\pmb{x}_{br}$ is the optimal strategic response (often termed as ``best response''). $u(\cdot)$ and $c(\cdot)$ represent designed utility and cost functions respectively. The utility function quantifies the benefit for agents in modifying their distribution from $\mathcal{D}$ to $\mathcal{D}(\theta)$, while the cost function captures the associated loss due to this shift. Once $\pmb{x}_{br}$ is determined, it can be used to recalibrate parameters $\theta$, initiating a new best response and leading to a feedback loop. An iterative approach, as demonstrated in~\cite{perdomo2020performative}, simulates this feedback loop to solve the strategic classification problem. This iterative process ultimately converges to a stable optimum where $\theta$ effectively classifies samples drawn from $\mathcal{D}(\theta)$.

A classic example is a bank employing a classifier to predict the creditworthiness of loan applicants. Individuals react to the bank's classifier by adjusting features (e.g., credit line utilization) strategically, invoking the performative distribution $\mathcal{D}(\theta)$ to influence favorable classification outcomes.

\section{Problem Formulation}
\label{sec:problem}

\par\noindent\textbf{Time-Series Forecasting.} 
Time-series forecasting involves predicting future values of one or more dependent time series based on historical data, potentially augmented with exogenous covariate features. We denote the target time series as $\bm{\mathrm{Y}}$ and its associated exogenous covariate features as $\bm{\mathrm{X}}$. At any time step $t$, time-series forecasting aims to predict $\bm{\mathrm{Y}}_t^H = [y_{t+1}, y_{t+2}, \ldots, y_{t+H}] \in \bm{\mathrm{Y}}$ using historical data $(\bm{\mathrm{X}}_t^L, \bm{\mathrm{Y}}_t^L)$, where $L$ represents the length of the historical data window, known as the \textit{lookback window}, and $H$ denotes the forecasting time steps, known as the \textit{horizon window}. Here, $\bm{\mathrm{X}}_t^L = [x_{t-L+1}, x_{t-L+2}, \ldots, x_{t}] \in \bm{\mathrm{X}}$ and $\bm{\mathrm{Y}}_t^L = [y_{t-L+1}, y_{t-L+2}, \ldots, y_{t}] \in \bm{\mathrm{Y}}$. For simplicity, we denote $\bm{\mathrm{Y}}^H = \{\bm{\mathrm{Y}}_t^H\}$ for $\forall t$ as the collection of horizon time-series of all time steps, similar for $\bm{\mathrm{Y}}^L$ and $\bm{\mathrm{X}}^L$. 
Conventional time-series forecasting aims to learn a model $\theta$ to obtain $\theta:(\bm{\mathrm{X}}^L, \bm{\mathrm{Y}}^L)\rightarrow \bm{\mathrm{Y}}^H$  that models the distribution $P(\bm{\mathrm{Y}}^H) = P(\bm{\mathrm{Y}}^H|\bm{\mathrm{Y}}^L)P(\bm{\mathrm{Y}}^L) + P(\bm{\mathrm{Y}}^H|\bm{\mathrm{X}}^L)P(\bm{\mathrm{X}}^L)$. This optimization is achieved through the empirical risk minimization (ERM):
\begin{equation}
    \theta^* = \arg\min_\theta \mathop{\mathbb{E}}_{\mathcal{Z}\sim \mathcal{D}} \mathcal{L} (\mathcal{Z};\theta)
    \label{eq:erm}
\end{equation}
where $\mathcal{D}$ denotes the distribution of training data, 
and $\mathcal{Z}$ are the samples drawn from the training data, i.e., $[(\bm{\mathrm{X}}_t^L, \bm{\mathrm{Y}}_t^L), \bm{\mathrm{Y}}_t^H]$ for a time step $t$.

\par\noindent\textbf{Performative Time-Series Forecasting.} In the performative setup of time-series forecasting, predictions can influence future data points and subsequent decision-making. Unlike traditional forecasting, which assumes data is drawn from a fixed distribution $\mathcal{D}$, performative forecasting considers samples drawn from a dynamic distribution that exogenous features can response to the forecasting model $\theta$. We refer to this evolving distribution as the performative distribution.   

Nevertheless, time-series forecasting features distinct differences compared to strategic classification problems. While predictive models in time series can influence actions, such as COVID-19 mortality and mobility forecasts as shown in Figure~\ref{fig:diag_intro}, these actions require a response time after predictions are made. This contrasts with the core assumption in strategic classification that have been broadly studied, where individuals can strategically modify their features before the predictive model is applied. That is, in the COVID-19 mortality and mobility forecasts example, individuals can only respond after the model predicts an outbreak, rather than preemptively altering their behavior before the prediction occurs.

Thus, unlike the hypothetical gaming in strategic classification, performative time-series forecasting involves predicting targets under a performative distribution that evolves over time. Instead of drawing samples from a dynamic distribution solely dependent on $\theta$, as in strategic classification, performative time-series forecasting requires optimizing under a co-evolving distribution that depends on both $\theta$ and time $t$, i.e., $\mathcal{D}(\theta, t)$. As such, we define the performative time-series forecasting as:
\par
\begin{figure}[t]
    \centering
    \begin{subfigure}{0.48\textwidth}
        \centering
        \includegraphics[width=0.99\textwidth]{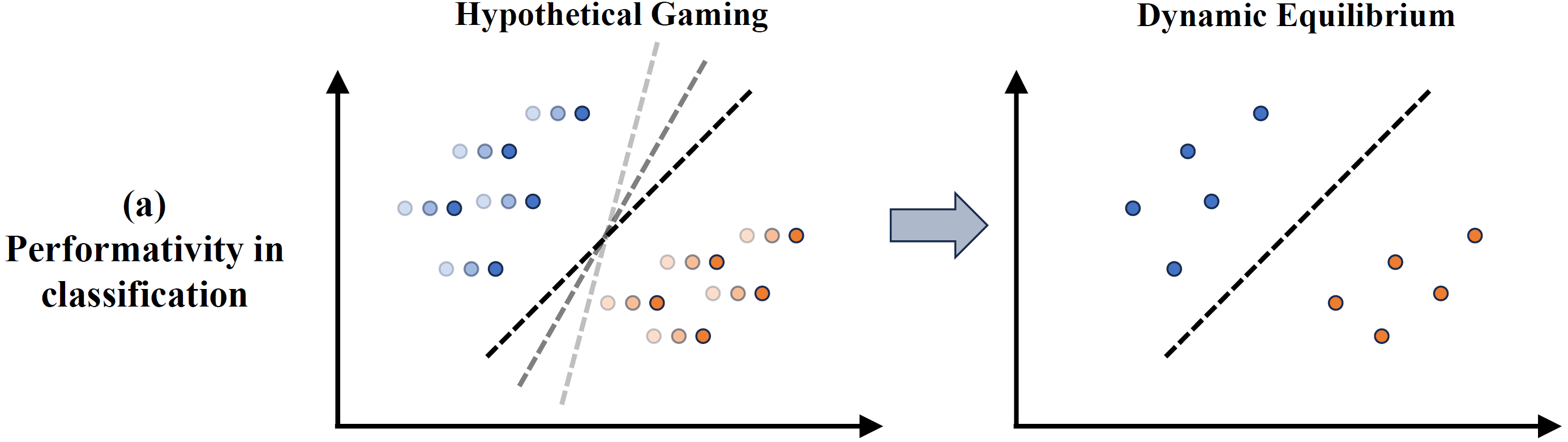}
        \label{fig:sub1}
    \end{subfigure}
    
    \vspace{1em} 
    
    \begin{subfigure}{0.48\textwidth}
        \centering
        \includegraphics[width=0.99\textwidth]{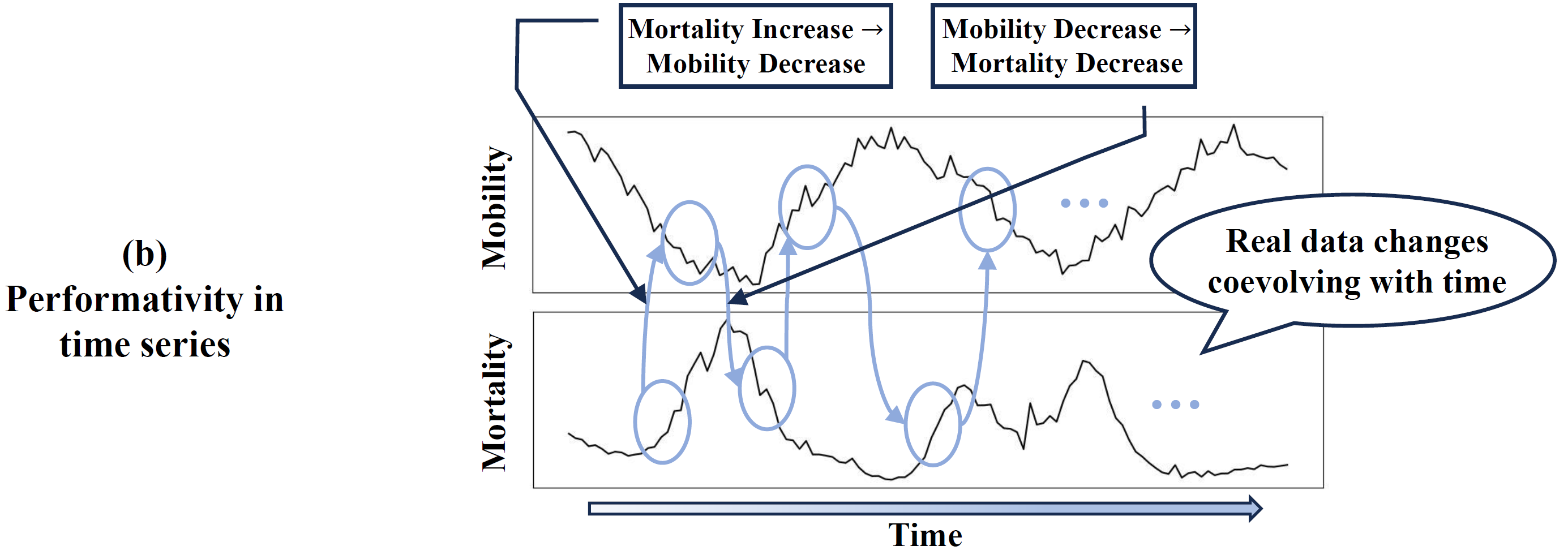}
        \label{fig:sub2}
    \end{subfigure}
    \caption{Comparison of performativity in strategic classification (synthetic example) and time-series forecasting (real COVID-19 example):  (a) Strategic classification involves a hypothetical interaction between features and decisions, aiming to find a dynamic equilibrium.  (b) Performative time-series forecasting, in contrast, deals with real data changes that evolve over time $t$ rather than hypothetical adaptations.  }
    \label{fig:diff}
\end{figure}

\noindent \textbf{Definition (\problem).} Performative time-series forecasting (\problem) involves making predictions under dynamic distribution shifts, where the distribution shift is induced by the model parameters $\theta$ and evolves over time $t$, by optimizing:
\begin{gather}
\label{eqn:obj_pets}
    \theta^*_{\problem} = \arg\min_\theta \mathop{\mathbb{E}}_{\mathcal{Z}(t) \sim \mathcal{D}(\theta, t)} \mathcal{L} (\mathcal{Z}(t);\theta)
\end{gather}
Here, we denote the samples $\mathcal{Z}(t)$ that is dependent on $t$, as $\mathcal{D}(\theta, t)$ evolves with $t$ and the samples are drawn from different time steps. We illustrate the difference between previous performative prediction studies and our performative time-series forecasting problems as shown in Figure~\ref{fig:diff}.

Since the distribution $\mathcal{D}(\theta, t)$ depends on both $\theta$ and $t$, and $\theta$ aims to minimize the empirical loss using data drawn from the dynamic distribution itself, a feedback loop—referred to as the performative feedback loop—naturally arises in \problem over time. This feedback loop can be described as the interactions between target time series and exogenous time series over time, using the multiplication of a series of conditional distributions, following the formulation:
\begin{equation}
\begin{aligned}
    P(\bm{\mathrm{Y}}_t^H) =& P(\bm{\mathrm{Y}}_t^H|\bm{\mathrm{X}}_{t+k_i}^L)P(\bm{\mathrm{X}}_{t+k_i}^L|\bm{\mathrm{Y}}_{t+k_{i-1}}^L) \ldots \\ 
    & P(\bm{\mathrm{Y}}_{t+k_2}^L|\bm{\mathrm{X}}_{t+k_1}^L) P(\bm{\mathrm{X}}_{t+k_1}^L|\bm{\mathrm{Y}}_t^L)P(\bm{\mathrm{Y}}_t^L), \\
    where& \:\: 0\leq k_1\leq k_2 \leq \ldots k_i \leq H \in \mathbb{N}
\end{aligned}
\label{eqn:loop}
\end{equation}

Intuitively, Equation~\ref{eqn:loop} illustrates how performativity influences the evolution of distributions over time, forming a performative feedback loop in time-series data. A decision made at time $t$, e.g. $P(\bm{\mathrm{Y}}_t^L)$, can trigger a response from individuals after a time delay $k_1$, leading to a shift in the distribution of $P(\bm{\mathrm{X}}_{t+k_1}^L)$. This altered distribution of exogenous features can subsequently impact back to the target time series after an additional delay $k_2$, resulting in a further distribution shift of $P(\bm{\mathrm{Y}}_{t+k_2}^L)$, so on and so forth. The Figure~\ref{fig:diff}.(b) showcases the feedback loop with the COVID-19 mortality and mobility example.

To address performative time-series forecasting, a natural solution involves both identifying the temporal correlations between the target and exogenous features—such as determining appropriate delays $k_1$ through $k_i$—and modeling the conditional distributions in Equation~\ref{eqn:loop} to accurately capture the performative feedback loop.

\par\noindent\textbf{Performative vs. Non-performative Features.} In practice, the performative distribution shift may not happen to all features. For instance, in COVID-19 forecasting, social mobility may shift in response to current forecasting results, while age distribution cannot shift whatever forecasting results are made. Following the preliminary setups in~\cite{perdomo2020performative}, we distinguish performative features and non-performative features in multivariate time series by defining: 

\par\noindent \textbf{Definition (Performative Feature).} The performative feature is a class of features whose distribution can actively shift as a response to the model predictions.

In general, multivariate time-series data may contain both performative and non-performative features.  \problem studies the distribution shift only over performative features.
Forecasting models of solving \problem hence take inputs from the concatenation shifted performative features and non-shifted non-performative features to conduct the target predictions.

\par\noindent\textbf{Performativity vs. Concept Drift.} Both performativity and concept drift delve into learning when the distributions of data change over time. While performativity specifically addresses the shift in the distribution of performative features due to predictive models, concept drift encompasses broader sources of distributional shift. To clarify, performativity studies the distribution shifts with specific shift orientation within the wider concept drift framework. Concept drift, in its broader scope, considers any arbitrary sources causing distributional changes.




\section{Methodology}
\label{sec:method}



While performativity plays a role in both strategic classification and time-series forecasting, solving performative time-series forecasting presents unique challenges. First, as shown in the comparison between Equation~\ref{eqn:perf} and Equation~\ref{eqn:obj_pets}, \problem involves a performative distribution shift that is not only induced by $\theta$ but also evolves with time $t$. This temporal dependent dynamic violates the fundamental assumption of strategic classification, making existing frameworks unsuitable for \problem. Second, the temporally dependent feedback loop, as shown in Equation~\ref{eqn:loop}, involves multiple interactions between target and exogenous features, making it particularly challenging to explicitly anticipate the distribution shift by modeling all associated conditional distributions.

\par\noindent \textbf{Methodology Overview.} 
Given the difficulty of both adapting existing frameworks and modeling all distribution shifts. We thereby propose an alternative shortcut to address \problem. On realizing the difficulty explicitly modeling all conditional distributions as in Equation~\ref{eqn:loop}, we propose a shortcut that models:
\begin{equation}
\begin{aligned}
    \mathrm{P}(\bm{\mathrm{Y}}^H_t) &= \mathrm{P}(\bm{\mathrm{Y}}_t^H|\bm{\mathrm{X}}^{L}_{t+k_i}) \allowbreak \mathrm{P}(\bm{\mathrm{X}}^{L}_{t+k_i}|\bm{\mathrm{X}}^{L}_t) \allowbreak \mathrm{P}(\bm{\mathrm{X}}_t^L)
\end{aligned}
\label{eq:sim}
\end{equation}
The intuition behind Equation~\ref{eq:sim} is to simplify the complex performative feedback loop from Equation~\ref{eqn:loop} into a two-stage forecasting process: an autoregressive process $\mathrm{P}(\bm{\mathrm{X}}^{L}_{t+k_i}|\bm{\mathrm{X}}^{L}_t)$ and a forecasting process $\mathrm{P}(\bm{\mathrm{Y}}_t^H|\bm{\mathrm{X}}^{L}_{t+k_i})$.  
The intuition of this simplification is in twofold: First, since the ultimate objective of time-series forecasting is to accurately predict the target $\mathrm{P}(\bm{\mathrm{Y}}^H_t)$, the most critical task is to model the deterministic conditional distribution $P(\bm{\mathrm{Y}}_t^H|\bm{\mathrm{X}}_{t+k_i}^L)$, given that $\bm{\mathrm{X}}_{t+k_i}^L$ directly influences $\bm{\mathrm{Y}}^H_t$ in the performative feedback loop. Second, while modeling $P(\bm{\mathrm{Y}}_t^H|\bm{\mathrm{X}}_{t+k_i}^L)$ can be achieved by conventional time-series forecasting models, it also presents two subsequent challenges: obtaining $\bm{\mathrm{X}}_{t+k_i}^L$ and estimating it effectively. To estimate, a natural approach for estimating $\bm{\mathrm{X}}_{t+k_i}^L$ is through an autoregressive process by modeling $\mathrm{P}(\bm{\mathrm{X}}^{L}_{t+k_i}|\bm{\mathrm{X}}^{L}_t)$. To determine $\bm{\mathrm{X}}_{t+k_i}^L$, we propose identifying the lagging effects of $\bm{\mathrm{X}}$ on $\bm{\mathrm{Y}}$ by maximizing their similarity over time. By finding this lag time, we determine $k_i$ and thus obtain $\bm{\mathrm{X}}_{t+k_i}^L$. For more straightforward intuitivity, we refer to $k_i$ as the delay $\tau$ and define $\bm{\mathrm{X}}_{t+k_i}^L$ as the ``\textbf{delayed response}'', denoted as $\bm{\mathrm{X}}^{DR} = \{\bm{\mathrm{X}}_{t+\tau}^L\}$.

By simplifying the performative feedback loop into this two-stage forecasting problem, we develop a more practical approach compared to directly solving Equation~\ref{eqn:loop}. For clearer notation and more intuitive understanding, we reformulate the two-step process in Equation~\ref{eq:sim} using the ``delayed response'' notation: The first step estimates the ``delayed response'' $\bm{\mathrm{X}}^{DR}$ from historical data $\bm{\mathrm{X}}^{L}$, while the second step forecasts the target $\bm{\mathrm{Y}}^{H}$ using the estimated $\bm{\mathrm{X}}^{DR}$. This is equivalent to:
\begin{equation}
\begin{aligned}
    \mathrm{P}(\bm{\mathrm{Y}}^H) &= \mathrm{P}(\bm{\mathrm{Y}}^H|\bm{\mathrm{X}}^{DR}) \allowbreak \mathrm{P}(\bm{\mathrm{X}}^{DR}|\bm{\mathrm{X}}^{L}) \allowbreak \mathrm{P}(\bm{\mathrm{X}}^L)
\end{aligned}
\label{eq:pets}
\end{equation}

\begin{figure}[t]
    \centering
    \hspace{0.0in}\includegraphics[width=0.48\textwidth]{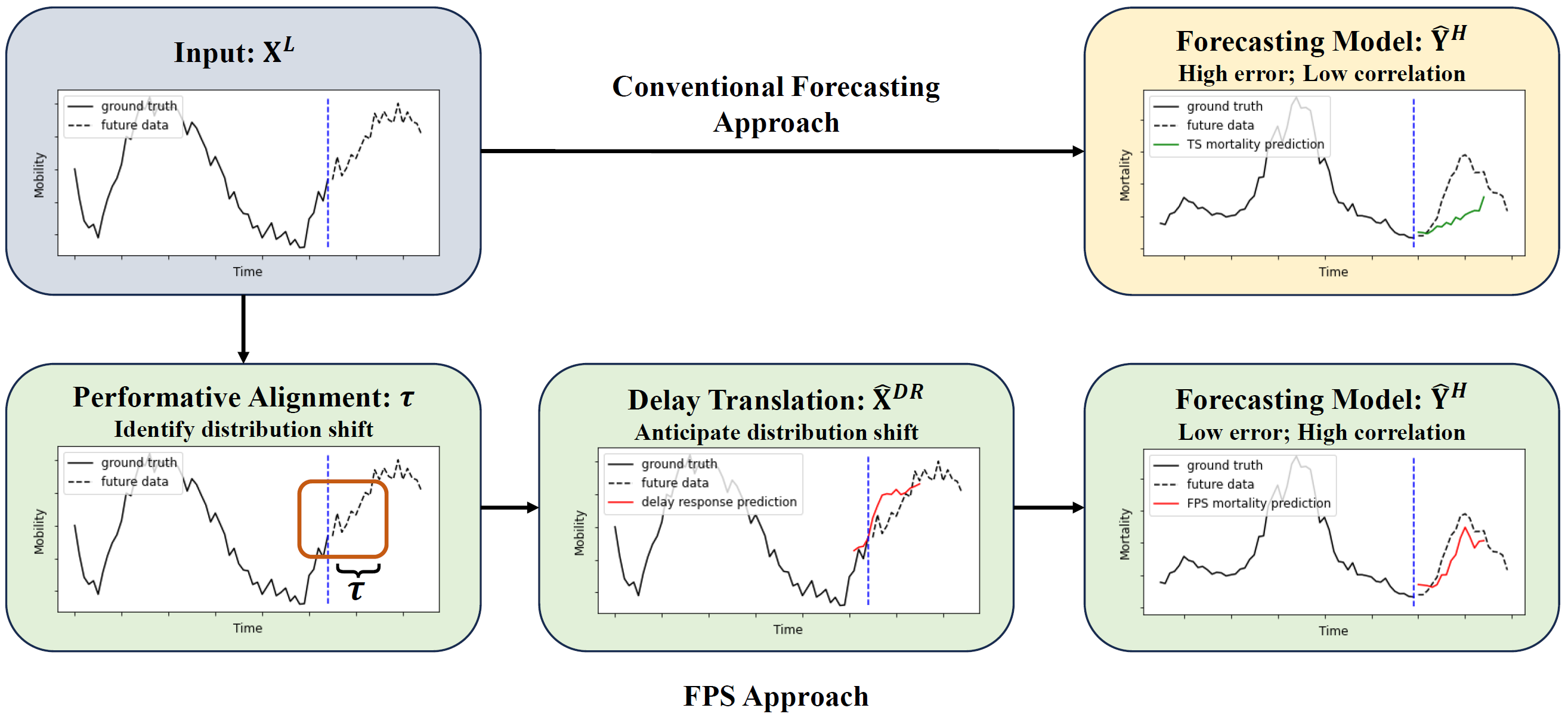}
    \caption{Comparison between the conventional forecasting approach and \solution. \solution utilizes a two-stage forecasting framework, which firstly estimates the delayed responses and then uses it to forecast the target variable (exampled on COVID-19 mobility and mortality data, trend changing of mobility is ahead of mortality).}
    \label{fig:diag_fps}
\end{figure}

Based on the above intuition and Equation~\ref{eq:pets}, we then propose \textit{\underline{F}eature} \textit{\underline{P}erformative} \textit{\underline{S}hifting} (\solution) to address \problem. \solution consists of three components, shown in Figure \ref{fig:diag_fps}: (1) Performative Time-Series Alignment, that determines the delay time $\tau$ such that $\bm{\mathrm{X}}^{DR}$ causes $\bm{\mathrm{Y}}^H$; (2) Delay Translation Module, that models the translation of $\mathrm{P}(\bm{\mathrm{X}}^{DR}|\bm{\mathrm{X}}^{L})$; and (3) Forecasting Module, that forecasts target $\bm{\mathrm{Y}}^H$ through modeling $\mathrm{P}(\bm{\mathrm{Y}}^H|\bm{\mathrm{X}}^{DR})$.

\subsection{Performative Time-Series Alignment}
\label{subsec:align}

To address the challenge of anticipating the distribution shift in \problem, it is crucial to determine the delay time $\tau$ where the shifted distribution $\bm{\mathrm{X}}^{DR}$ directly causes $\bm{\mathrm{Y}}^{H}$. Consider the COVID-19 forecasting example at a specific time step $t$: the performative feedback loop can be viewed as a two-step process. First, a prediction of reduced mortality can lead to increased mobility; second, this increased mobility can cause another outbreak of the pandemic. Viewing the prediction of the pandemic outbreak as the target $\bm{\mathrm{Y}}^{H}_{t}$, the increase in mobility is the direct cause of this target, making $\bm{\mathrm{X}}^{DR}_t \equiv \bm{\mathrm{X}}^{L}_{t+\tau}$ share a similar trend with the target series.

Thus, the key idea of Performative Time-Series Alignment is to determine the delay time $\tau$ by maximizing the similarity between the target series and a sliding window of the exogenous features on the full training data, defined as:
\begin{equation}
\begin{aligned}
    \tau = \arg\max_{i} &|\textrm{Similarity}([x_{i}, \ldots, x_{T-H+i}], \\ &[y_{H}, \ldots, y_{T}])|, \:\:\: i \in[0,H]
\label{eq:tau}
\end{aligned}
\end{equation}
The concept of assessing the similarity between delayed responses and target predictions can be extended to general time-series forecasting tasks. If an input feature positively correlates with the target prediction, a trend change in the delayed response will induce a similar trend change in the target prediction, maximizing the similarity. Conversely, if the input feature negatively correlates with the target prediction, a trend change in the delayed response will lead to an opposite trend change in the target prediction, minimizing the similarity while maximizing its absolute value. This alignment can be achieved by calculating similarity measures between input and target time series at various time shifts, similar to the widely-used Dynamic Time Warping technique~\cite{muller2007dynamic}. We employ cosine similarity, defined as $\textsc{CS}(\bm{\mathrm{X}},\bm{\mathrm{Y}}) = \frac{\bm{\mathrm{X}} \cdot \bm{\mathrm{Y}}}{|\bm{\mathrm{X}}||\bm{\mathrm{Y}}|}$, to measure the similarity between input performative features and target time series~\cite{nakamura2013shape}. When input performative features are multivariate, each dimension is independently aligned with the target. Different metrics can be used to obtain $\tau$, such as maximizing Pearson Correlation or minimizing Euclidean Distance~\cite{coakley2001alignment, kumar2010iterative}.

\subsection{Delay Translation Module}
\label{subsec:delay}

To address the challenge of unobserved future data for delayed responses $\bm{\mathrm{X}}^{DR}$ during testing as $t+\tau$ can possibly larger than $T$, we introduce the Delay Translation Module to model the translation $\mathrm{P}(\bm{\mathrm{X}}^{DR}|\bm{\mathrm{X}}^{L})$. This module accurately predicts the delayed response based on the present input feature, implicitly anticipating the distribution shift from $\mathcal{D}$ to $\mathcal{D}(\theta, t)$ in Equation \ref{eq:pets} in Equation~\ref{eqn:obj_pets}. Consequently, the task of anticipating the distribution shift can be reframed as an autoregressive problem, which has been well-established~\cite{hyndman2018forecasting}. The Delay Translation Module, denoted as $f_\tau$, addresses this autoregressive task using sequence-to-sequence mapping neural networks, characterizing better approximation capability and generalization ability. We employ Recurrent Neural Networks (RNNs), a common and simple approach for sequence predictions, which is optimized by minimizing the empirical mean squared error (MSE) risk:
\begin{equation}
\label{eq:loss_dt}
    \mathcal{L}_{DT} (f_\tau) = \textsc{MSE}(\:\bm{\mathrm{X}}^{DR}, \: \hat{\bm{\mathrm{X}}}\vphantom{\bm{\mathrm{X}}}^{DR} = f_\tau(\bm{\mathrm{X}}^L)\:)
\end{equation}

In practice, while this process may introduce additional errors, it generally captures trend changes in advance, providing sufficient support for target prediction. Meanwhile, although simple RNNs have consistently outperformed other models in empirical evaluations, more advanced architectures, such as DeepAR~\cite{salinas2020deepar} or Transformer-based models, can also be employed to further enhance performance in autoregressive delay translation tasks.

\subsection{Forecasting Module}
\label{subsec:forecast}

After obtaining the estimated delayed response $\hat{\bm{\mathrm{X}}}\vphantom{\bm{\mathrm{X}}}^{DR} = f_\tau(\bm{\mathrm{X}}^L)$ from the Delay Translation Module, the problem turns to a conventional time-series forecasting involving multivariate time-series data. The role of the forecasting module is to predict the targets $\bm{\mathrm{Y}}^H$ using $\hat{\bm{\mathrm{X}}}\vphantom{\bm{\mathrm{X}}}^{DR}$ along with non-performative features. For ease of reference, we represent $\hat{\bm{\mathrm{X}}}\vphantom{\bm{\mathrm{X}}}^{DR}$ as the concatenation of the estimated delayed response and non-performative features. Recognizing the delayed responses $\bm{\mathrm{X}}^{DR}$ are often unavailable as future time series, the forecasting module employs the estimated $\hat{\bm{\mathrm{X}}}\vphantom{\bm{\mathrm{X}}}^{DR}$ to forecast the targets during training, validation, and testing.

The objective of the time-series forecasting module, denoted as $g_\tau$, is to minimize the risk of the designed loss function, which is usually mean squared error (MSE):
\begin{equation}
\label{eq:loss_ts}
    \mathcal{L}_{TS} (g_\tau) = \textsc{MSE}(\:\bm{\mathrm{Y}}^H, \: \hat{\bm{\mathrm{Y}}}\vphantom{\bm{\mathrm{Y}}}^{H}=g_\tau(\hat{\bm{\mathrm{X}}}\vphantom{\bm{\mathrm{X}}}^{DR}, \bm{\mathrm{Y}}^L)\:)
\end{equation}
The key difference between the forecasting module in \solution and conventional time-series forecasting is that \solution employs estimated delayed responses $\hat{\bm{\mathrm{X}}}\vphantom{\bm{\mathrm{X}}}^{DR}$ in the horizon window, which models the direct cause to target predictions. In contrast, conventional time-series forecasting simply uses $\bm{\mathrm{X}}^L$ from the lookback window, which does not consider performativity and may model incorrect causal effects among time steps, as $\bm{\mathrm{X}}^L$ may not directly cause $\bm{\mathrm{Y}}^H$ at many cases.

The choice of the forecasting module is independent of the delay translation module, allowing for the flexible selection of any state-of-the-art time-series forecasting model. The loss objective for the forecasting module may include additional penalties, depending on the chosen model.


\subsection{Feature Performative-Shifting - \solution}

We integrate all the aforementioned modules to formulate the solution for \problem: \textit{Feature Performative-Shifting} (\solution). In \solution, we first determine the time delay $\tau$ through performative time-series alignment. We then co-optimize the delay translation module $f_{\tau}(\cdot)$ and the forecasting module $g_{\tau}(\cdot)$. The complete model of \solution is denoted as $\theta_\solution = g_{\tau}(f_{\tau}(\cdot))$. The joint optimization of the delay translation module and the forecasting module is accomplished by minimizing the combined loss:
\begin{gather}
\label{eqn:obj_fps}
    \mathcal{L}_\solution = \lambda_1 \mathcal{L}_{DT}(f_{\tau}) + \lambda_2 \mathcal{L}_{TS}(g_{\tau})
\end{gather}
Here, $\lambda_1$ and $\lambda_2$ act as hyperparameters, allowing the weighting of loss terms without loss of generality; both can be set to 1 with proven effectiveness in practice. The detailed training process of \solution is outlined in Algorithm \ref{alg:fps}. Intuitively, \solution is a two-stage forward algorithm: first, it predicts the intermediate target—the delayed responses due to the performative distribution shift—and then forecasts the desired target using the predicted delayed responses.

While \solution concurrently trains both the delay translation and forecasting modules, it is also viable to train these two modules independently in practice. Specifically, the delay translation module can be pre-trained or updated with lower frequency, providing \solution with enhanced efficiency and flexibility in real-time forecasting tasks.

\begin{algorithm}
	\caption{Feature Performative-Shifting (\solution)}\label{alg:fps}
	\begin{algorithmic}[1]
    \State \textbf{Input:} time-series dataset $\mathcal{Z}=\{\bm{\mathrm{X}}, \bm{\mathrm{Y}}\}$, max time $T$, lookback $L$, horizon $H$, epoch $E$, initial model $\theta_0$
    \State \textbf{Output:}  $\theta_\solution$ for solving \problem
    \State Initialize $\tau = 0$,\: $\textsc{CS}_{max} = 0$
	\For {$i=0,\ldots,H$} 
		\State Compute $\textsc{CS}([x_{i}, \ldots, x_{T-H+i}],[y_{H}, \ldots, y_{T}])$
		\If {$|\textsc{CS}| \geq \textsc{CS}_{max}$}
            \State $\tau = i$,\: $\textsc{CS}_{max} = \textsc{CS}$ 
        \EndIf
    \EndFor
    \State Sample training data $\bm{\mathrm{X}}^L$, $\bm{\mathrm{X}}^{DR}=\{\bm{\mathrm{X}}^L_{t+\tau}\}$, $\bm{\mathrm{Y}}^L$, $\bm{\mathrm{Y}}^H$
    \For{epoch $=1, \cdots, E$}
        \State Compute $\hat{\bm{\mathrm{X}}}\vphantom{\bm{\mathrm{X}}}^{DR} = f_\tau(\bm{\mathrm{X}}^L)$ 
        \State Compute $\hat{\bm{\mathrm{Y}}}\vphantom{\bm{\mathrm{Y}}}^{H}=g_\tau(\hat{\bm{\mathrm{X}}}\vphantom{\bm{\mathrm{X}}}^{DR}, \bm{\mathrm{Y}}^L)$ 
        \State \parbox[t]{\dimexpr\linewidth-\algorithmicindent}{Compute $\mathcal{L}_\solution$ and update $\theta_i = g_{\tau}(f_{\tau}(\cdot))$ with gradient-based method.}
    \EndFor
    \State \textbf{return} $\theta_\solution = \theta_E$
	\end{algorithmic} 
\end{algorithm} 

During evaluation, we assume that the delay time $\tau$ determined during training also holds for the validation and testing data. Based on this assumption, \solution processes the input features through the delay translation module and subsequently through the forecasting module. For given inputs, the final forecasting out of \solution aligns same with conventional time-series methods optimized through ERM.


\par\noindent\textbf{Analysis of \solution: Potential for tightening PAC bounds.} To better understand why \solution may improve forecasting accuracy, we propose the following two propositions to explain the performance disparities between conventional forecasting approaches and \solution:
\begin{proposition}
\label{prop1}
    Assume $\bm{\mathrm{X}}^{DR}$ is known for both training and testing, and $\bm{\mathrm{Y}}^{H}$ is known for training yet not in testing,
    employing $\bm{\mathrm{X}}^{DR}$ to forecast $\bm{\mathrm{Y}}^{H}$ yields a tighter PAC bound compared to using original input sequences. 
\end{proposition}
\begin{proposition}
\label{prop2}
    Assume both $\bm{\mathrm{X}}^{DR}$ and $\bm{\mathrm{Y}}^{H}$ are known for training yet not testing, which requires the delay translation module to forecast the estimated delayed responses $\hat{\bm{\mathrm{X}}}\vphantom{\bm{\mathrm{X}}}^{DR}$ during the testing stage, \solution reduces empirical loss by increasing model complexity of the PAC bound compared to conventional forecasting methods.
\end{proposition}
\par \noindent
The overall idea for proving the propositions is to show that the PAC bound is upper bounded by three terms: the empirical loss, the model complexity, and the non-vanishing errors. This is achieved by generalizing and adapting the theorems from previous studies—\cite{mohri2008rademacher} for stationary cases and~\cite{kuznetsov2017generalization} for non-stationary cases. We then show that forecasting with $\bm{\mathrm{X}}^{DR}$ results in a lower empirical error but a higher model complexity, as it involves a two-stage forecasting method. The detailed proof is provided in Appendix I.  

Intuitively, Proposition \ref{prop1} describes an ideal setting where $\bm{\mathrm{X}}^{DR}$, which contains partial future values of exogenous features, is assumed to be known in advance. In this case, forecasting with $\bm{\mathrm{X}}^{DR}$ results in a lower empirical error, while preserving the same model complexity as forecasting with $\bm{\mathrm{X}}^{L}$. This leads to a tighter PAC bound when forecasting with $\bm{\mathrm{X}}^{DR}$, this is further shown empically through Table~\ref{tbl:ideal}.

Proposition \ref{prop2} describes the practical setting, where the future values of $\bm{\mathrm{X}}^{DR}$ are unknown. In this case,  while \solution achieves lower empirical error by forecasting with $\hat{\bm{\mathrm{X}}}\vphantom{\bm{\mathrm{X}}}^{DR}$, the introduction of the delayed translation module increases model complexity, which introduces a trade-off between lower empircal loss and higher model complexity in the PAC bound. 

Indeed, quantatively determining the trade-off between the increase in model complexity and the reduction in empirical error is non-trivial. However, empirical evaluations demonstrate that the reduction in empirical loss outweighs the increase in model complexity in practice. As a result, leveraging forecasted delayed responses, even within \solution, proves beneficial for forecasting performance. This intuition is further shown by Table~\ref{tbl:ts} and Table~\ref{tbl:rt}.  

\section{Experiments}
\label{sec:exp}

\begin{table*}[t]
\bgroup
\def\arraystretch{1.05}
\caption{Performance comparison between baselines and \solution for standard time-series forecasting. \solution shows better performance in 19 out of 20 evaluations compared to all baselines across various forecasting models and evaluation metrics.}
\resizebox{0.9\textwidth}{!}{%
\begin{tabular}{c|cccccccccc}
\hline\hline
Dataset  & \multicolumn{10}{c}{\texttt{\textbf{covid}}} \\
Model    & \multicolumn{2}{c|}{RNN} & \multicolumn{2}{c|}{LSTNet} & \multicolumn{2}{c|}{Transformer} & \multicolumn{2}{c|}{Informer} & \multicolumn{2}{c}{Autoformer} \\
Method \textbf{\textbackslash} \! Metric & NMAE & \multicolumn{1}{c|}{NRMSE} & NMAE & \multicolumn{1}{c|}{NRMSE} & NMAE & \multicolumn{1}{c|}{NRMSE} & NMAE & \multicolumn{1}{c|}{NRMSE} & NMAE & NRMSE \\ \hline
ERM      & 0.288 & \multicolumn{1}{c|}{0.491} & 0.223 & \multicolumn{1}{c|}{0.479} & 0.249 & \multicolumn{1}{c|}{0.491} & 0.305 & \multicolumn{1}{c|}{0.626} & 0.810 & 1.772 \\
GroupDRO & 0.263 & \multicolumn{1}{c|}{0.424} & 0.233 & \multicolumn{1}{c|}{0.381} & 0.409 & \multicolumn{1}{c|}{0.765} & 0.352 & \multicolumn{1}{c|}{0.659} & 0.524 & 1.008 \\
IRM      & 0.303 & \multicolumn{1}{c|}{0.501} & 0.271 & \multicolumn{1}{c|}{0.487} & 0.548 & \multicolumn{1}{c|}{1.048} & 0.517 & \multicolumn{1}{c|}{0.995} & 0.449 & 0.852 \\
RevIN   & 0.337 & \multicolumn{1}{c|}{0.669} & 0.379 & \multicolumn{1}{c|}{0.738} & 0.478 & \multicolumn{1}{c|}{0.898} & 0.394 & \multicolumn{1}{c|}{0.744} & 0.521 & 1.011 \\
Dish-TS  & 0.388 & \multicolumn{1}{c|}{0.736} & 0.291 & \multicolumn{1}{c|}{0.516} & 0.322 & \multicolumn{1}{c|}{0.530} & 0.338 & \multicolumn{1}{c|}{0.569} & 0.442 & 0.821 \\
\solution (Ours) & \textbf{0.194} & \multicolumn{1}{c|}{\textbf{0.321}} & \textbf{0.185} & \multicolumn{1}{c|}{\textbf{0.311}} & \textbf{0.179} & \multicolumn{1}{c|}{\textbf{0.259}} & \textbf{0.231} & \multicolumn{1}{c|}{\textbf{0.396}} & \textbf{0.370} & \textbf{0.678} \\ \hline\hline
Dataset  & \multicolumn{10}{c}{\texttt{\textbf{traffic}}} \\
Model    & \multicolumn{2}{c|}{RNN} & \multicolumn{2}{c|}{LSTNet} & \multicolumn{2}{c|}{Transformer} & \multicolumn{2}{c|}{Informer} & \multicolumn{2}{c}{Autoformer} \\
Method \textbf{\textbackslash} \! Metric & NMAE & \multicolumn{1}{c|}{NRMSE} & NMAE & \multicolumn{1}{c|}{NRMSE} & NMAE & \multicolumn{1}{c|}{NRMSE} & NMAE & \multicolumn{1}{c|}{NRMSE} & NMAE & NRMSE \\ \hline
ERM      & 0.141 & \multicolumn{1}{c|}{0.208} & 0.135 & \multicolumn{1}{c|}{0.196} & 0.145 & \multicolumn{1}{c|}{0.216} & 0.141 & \multicolumn{1}{c|}{0.198} & 0.166 & 0.217 \\
GroupDRO & 0.132 & \multicolumn{1}{c|}{0.194} & 0.127 & \multicolumn{1}{c|}{0.181} & 0.151 & \multicolumn{1}{c|}{0.216} & 0.130 & \multicolumn{1}{c|}{0.187} & 0.179 & 0.243 \\
IRM      & 0.135 & \multicolumn{1}{c|}{0.196} & 0.125 & \multicolumn{1}{c|}{0.175} & 0.147 & \multicolumn{1}{c|}{0.218} & 0.131 & \multicolumn{1}{c|}{0.184} & 0.191 & 0.259 \\
RevIN   & 0.166 & \multicolumn{1}{c|}{0.232} & 0.163 & \multicolumn{1}{c|}{0.226} & 0.164 & \multicolumn{1}{c|}{0.225} & 0.170 & \multicolumn{1}{c|}{0.234} & 0.196 & 0.266 \\
Dish-TS  & 0.135 & \multicolumn{1}{c|}{\textbf{0.185}} & 0.148 & \multicolumn{1}{c|}{0.198} & 0.135 & \multicolumn{1}{c|}{0.187} & 0.140 & \multicolumn{1}{c|}{0.190} & 0.172 & 0.222 \\
\solution (Ours) & \textbf{0.131} & \multicolumn{1}{c|}{\underline{0.194}} & \textbf{0.119} & \multicolumn{1}{c|}{\textbf{0.171}} & \textbf{0.129} & \multicolumn{1}{c|}{\textbf{0.179}} & \textbf{0.114} & \multicolumn{1}{c|}{\textbf{0.166}} & \textbf{0.141} & \textbf{0.194} \\ \hline\hline
\end{tabular}
}
\label{tbl:ts}
\egroup
\end{table*}

\subsection{Setup}
\par\noindent \textbf{Datasets.} Our empirical evaluations use the COVID-19~\cite{rodriguez2022einns} and METR-LA traffic~\cite{li2018dcrnn_traffic} datasets, both of which offer clearly identifiable performative features, such as mobility data in the COVID-19 dataset and traffic volumes from neighboring sensors in the METR-LA dataset. Our forecasting tasks are to predict national and state-wise weekly mortality for the next 8 weeks using the COVID-19 dataset and to forecast traffic volume every 5 minutes for the next 2 hours using the METR-LA dataset.

We evaluate \solution under two settings: standard setting~\cite{zhou2021informer} and real-time setting~\cite{rodriguez2022einns}. In standard setting, the time series is divided into 60\% for training, 20\% for validation, and 20\% for testing. For real-time setting, we make weekly re-training and predictions from October 2021 to March 2022 for the COVID-19 dataset and hourly predictions throughout the day for the METR-LA dataset. The first 20\% of real-time steps are used for validation. This setup is consistent with prior research~\cite{rodriguez2022einns, li2018dcrnn_traffic}. Additional experiment details are available in Appendix II.

\par\noindent\textbf{Baselines.} Our primary objective is to showcase the advantages of \solution compared to existing concept drift and non-stationary approaches in addressing \problem. We include five baselines (referred to as ‘Method’ in the result tables): Empirical Risk Minimization (ERM) as defined in Equation \ref{eq:erm}, two general concept drift methods, GroupDRO~\cite{sagawa2019distributionally} and IRM~\cite{arjovsky2019invariant}, and two non-stationary methods specifically designed for time-series forecasting, RevIN~\cite{kim2021reversible} and Dish-TS~\cite{fan2023dish}. We exclude other concept drift baselines, such as SD~\cite{pezeshki2021gradient} and IB-ERM~\cite{ahuja2021invariance}, as they do not show significant empirical benefits over ERM for time-series forecasting tasks. 

Since all methods are model-agnostic frameworks, we include five forecasting models (referred to as ‘Model’ in the result tables): RNNs, LSTNet~\cite{lai2018modeling}, Transformer~\cite{vaswani2017attention}, Informer~\cite{zhou2021informer}, and Autoformer~\cite{wu2021autoformer}. This inclusion demonstrates that the benefits of \solution are consistent and independent of the specific forecasting model employed.


\par\noindent \textbf{Evaluation.} 
We measure performance disparities using Normalized Root Mean Squared Error (NRMSE) and Normalized Mean Absolute Error (NMAE) for standard time-series forecasting tasks, and additionally Pearson Correlation (PC) for real-time forecasting tasks. NRMSE and NMAE measure forecasting accuracy, while PC evaluates the models' abilities to capture trends. The detailed metrics' formulations are:
\begin{gather*}
    \textsc{NMAE} = \frac{\sum_{N,H}|y_{n,h} - \hat{y}_{n,h}|}{\sum_{N,H}|y_{n,h}|}   \\
    \textsc{NRMSE} = \frac{\sqrt{\frac{1}{NH}\sum_{N,H}(y_{n,h} - \hat{y}_{n,h})^2}}{\frac{1}{NH}\sum_{N,H}|y_{n,h}|} \\
    \textsc{PC} = \frac{\sum_{H}(y_h - \bar{y}_h)(\hat{y}_h - \bar{\hat{y}}_h)}{\sqrt{\sum_H (y_h - \bar{y}_h)^2(\hat{y}_h - \bar{\hat{y}}_h)^2}} 
\end{gather*}
Here, $N$ and $n$ denote the total number and the index of testing sequences, respectively, while $H$ and $h$ represent the horizon length and the time steps within each test sequence.


\par\noindent \textbf{Reproducibility.} All models are trained on NVIDIA Tesla V100 GPUs. All data and code are available at: \url{https://github.com/AdityaLab/pets}.

\subsection{Results}

\par\noindent \textbf{Standard Time-Series Forecasting.} 
We conducted experiments on standard time-series forecasting tasks, dividing the data into 60\% for training, 20\% for validation, and 20\% for testing. The evaluation compares the conventional approach (Empirical Risk Minimization, ERM) with two concept drift methods (GroupDRO and IRM), two non-stationary methods (RevIN and Dish-TS), and \solution. The comprehensive results, detailed in Table\ref{tbl:ts}, led to three key conclusions:

\par\noindent(1). \textit{Introducing performativity to ERM enhances forecasting accuracy}. \solution can be conceptualized as a two-stage process. In the first stage, it anticipates performative distribution shifts by estimating delayed responses, and in the second stage, it forecasts the target based on these estimations. Both stages of \solution employ ERM, making \solution a two-stage ERM with performativity introduced. Evaluation results consistently demonstrate an improvement in \solution over ERM, with average error reductions of 37.1\% and 12.8\% for COVID-19 and traffic forecasting tasks, respectively. This improvement highlights the advantage of introducing performativity to ERM in addressing \problem with \solution.

\par\noindent(2). \textit{\solution better addresses distribution shifts in scenarios with performativity compared to other general distribution shift methods.} The key difference between \solution and existing distribution drift methods lies in the manner of handling distribution shifts. While general distribution drift methods focus on mitigating the effects of distribution shifts by learning environment-invariant patterns or utilizing instance normalization, \solution anticipates distribution shifts due to performativity, thereby enhancing forecasting outcomes. Evaluation results demonstrate that \solution consistently achieves lower forecasting errors than popular concept drift methods on COVID-19 and traffic datasets, whose performativity impacts are clearly presented, outperforming in 19 out of 20 comparisons with time-series non-stationary methods. These findings highlight the benefits of anticipating the impact of distribution shifts in time-series forecasting tasks, rather than merely mitigating or normalizing them. Notably, time-series-specific distribution shift methods, such as RevIN, do not significantly improve short-term forecasting performance, as their instance normalization processes limit potential trend changes during prediction. A detailed analysis, including case studies, can be found in Appendix III.

\par\noindent(3). \textit{Empirical evaluation of \solution validates theoretical analysis.} Recall that Proposition~\ref{prop2} introduces the trade-off of \solution to the PAC bound, where \solution yields lower empirical training error but higher model complexity. Our empirical evaluations confirm that the reduction in empirical loss outweighs the increase in model complexity when comparing ERM and \solution, thus leading to lower testing errors. Additionally, experiments directly utilizing $\bm{\mathrm{X}}^{DR}$ (i.e., future time-series from the horizon window) for traffic predictions show a 56.3\% reduction in average forecasting errors except when using Autoformer (Table~\ref{tbl:ideal} in Appendix III). These additional results validate Proposition~\ref{prop1}, demonstrating that forecasting with $\bm{\mathrm{X}}^{DR}$ yields a lower error bound.

\begin{table*}[!t]
\centering
\bgroup
\caption{Performance comparison between ERM and \solution for real-time forecasting tasks. \solution uniformly outperforms ERM on both datasets and all metrics ($\Downarrow$ means lower is better, $\Uparrow$ means higher is better)}
\def\arraystretch{1.05}
\begin{tabular}{cc|ccc|ccc}
\hline \hline
& & \multicolumn{3}{c}{\textbf{\texttt{covid}}} & \multicolumn{3}{c}{\textbf{\texttt{traffic}}} \\
Model                       & Method & NMAE($\Downarrow$) & NRMSE($\Downarrow$) & PC($\Uparrow$) & NMAE($\Downarrow$) & NRMSE($\Downarrow$) & PC($\Uparrow$)    \\ \hline
\multirow{2}{*}{RNNs}          & ERM & 0.454 & 1.061 & 0.526 & 0.168 & 0.235 & 0.272       \\
                                        & \solution & \textbf{0.405} & \textbf{0.877} & \textbf{0.661} & \textbf{0.138} & \textbf{0.198} & \textbf{0.562}        \\ \hline
\multirow{2}{*}{LSTNet}        & ERM & 0.324 & 0.597 & 0.656 & 0.177 & 0.251 & 0.550       \\
                                        & \solution & \textbf{0.261} & \textbf{0.570} & \textbf{0.818} & \textbf{0.130} & \textbf{0.190} & \textbf{0.723}        \\ \hline
\multirow{2}{*}{Transformer}   & ERM & 0.551 & 1.048 & 0.159 & 0.202 & 0.277 & 0.257       \\
                                        & \solution & \textbf{0.442} & \textbf{0.898} & \textbf{0.401} & \textbf{0.133} & \textbf{0.202} & \textbf{0.571}       \\ \hline
\multirow{2}{*}{Informer}      & ERM & 0.392 & 0.958 & 0.693 & 0.142 & 0.201 & 0.487       \\
                                        & \solution & \textbf{0.340} & \textbf{0.883} & \textbf{0.791} & \textbf{0.138} & \textbf{0.200} & \textbf{0.651}        \\ \hline
\multirow{2}{*}{Autoformer}      & ERM & 0.435 & 0.833 & 0.535 & 0.183 & 0.234 & 0.339       \\
                                        & \solution & \textbf{0.312} & \textbf{0.568} & \textbf{0.697} & \textbf{0.140} & \textbf{0.194} & \textbf{0.482}        \\ \hline \hline
\end{tabular}
\label{tbl:rt}
\egroup
\end{table*}

\par\noindent \textbf{Real-Time Forecasting.} 
We explore real-time forecasting, a critical task with practical applications such as CDC requiring weekly updates on COVID-19 predictions. Real-time forecasting involves sequentially predicting imminent values or events as the most current data becomes available. Practical real-time data often exhibit dynamic patterns, including seasonal variations, interventions, sudden anomalies, and trend changes, posing distribution shift challenges and necessitating adaptable forecasting models capable of adjusting to evolving patterns or abrupt shifts in data.

To evaluate the efficacy of \solution in addressing \problem within real-time forecasting, we compare its performance with that of ERM in practical real-time scenarios, as detailed in Table \ref{tbl:rt}. We include Pearson Correlation to highlight differences in trend correlations between \solution and the conventional forecasting method. Concept drift and non-stationary methods are excluded for two reasons: (a) these methods are designed to mitigate distribution shifts for stable predictions, which contrasts with capturing dynamic patterns and rapid trend changes in real-time forecasting, and (b) we have demonstrated that \solution outperforms distribution shift baselines in earlier evaluations.

The evaluation demonstrates that \solution consistently outperforms conventional approaches across all testing metrics. Specifically, \solution achieves an average 16.3\% error reduction and a 47.9\% improvement in correlations for COVID-19 forecasting, and an average 18.9\% error reduction and a 67.1\% improvement in correlations for traffic forecasting. The importance of real-time forecasting, as evidenced by the CDC's requirement for timely updates on COVID-19 and flu predictions, underscores the value of \solution. Its adaptive nature makes it a crucial tool for decision-makers who rely on precise and timely forecasts, particularly in domains where timely insights have a significant impact.

\par\noindent\textbf{Case Study: Capturing Trend Changes with Minimal Delay.} To illustrate the benefits of \solution in anticipating performative distribution shifts and effectively learning dynamic patterns, we present a case study on real-time forecasting of COVID-19 trends. This case study uses data up to Week 2022-03 to predict national mortality rates for Weeks 2022-04 through 2022-11. Leading up to Week 2022-03, the national mortality rate had risen steadily due to the Omicron outbreak beginning in December 2021. In response to the foreseen outbreak, public mobility decreased, leading to a subsequent decline in the mortality rate starting from Week 2022-04. This case study highlights how \solution captures the performative shifts associated with the Omicron outbreak, public response, and the subsequent decline in the pandemic.

\begin{figure}[h]
    \centering
    \includegraphics[width=0.36\textwidth]{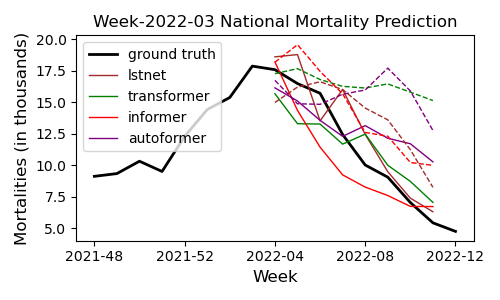}
    \caption{Case Study: Plots of national mortality prediction at the trend turning point, with and without \solution. (Solid and dashed plots for with and without \solution, respectively)}
    \label{fig:case}
\end{figure}

The case study showcases real-time forecasting results of with and without \solution, with qualitative and quantitative results shown in Figure~\ref{fig:case} and Table~\ref{tbl:case}. Qualitative, \solution consistently provides a more accurate prediction of trend changes compared to ERM. Specifically, \solution captures steady and precise decreasing trends, while ERM exhibits delayed decreases. Quantitatively, predictions derived from the \solution model consistently exhibit lower forecasting errors and higher correlations, highlighting its effectiveness in capturing trend changes with minimal time delay. These findings underscore \solution's advantages in detecting rapid and abrupt shifts in trends.

\begin{table}[h]
\centering
\caption{Case Study: Comparison between conventional forecasting models and \solution for real-time forecasting at Week 2022-03. Our method \solution uniformly outperforms conventional forecasting models on all metrics.}
\bgroup
\def\arraystretch{1.00}
\begin{tabular}{cc|ccc}
\hline\hline
Model                        & Method & NMAE($\Downarrow$) & NRMSE($\Downarrow$) & PC($\Uparrow$) \\ \hline
\multirow{2}{*}{LSTNet}      & ERM & 0.250 & 0.281 & 0.845 \\
                             & \solution & \textbf{0.140} & \textbf{0.167} & \textbf{0.932}\\ \hline
\multirow{2}{*}{Transformer} & ERM & 0.409 & 0.505 & 0.922\\
                             & \solution & \textbf{0.160} & \textbf{0.172} & \textbf{0.939}\\ \hline
\multirow{2}{*}{Informer}    & ERM & 0.236 & 0.255 & 0.911 \\
                             & \solution & \textbf{0.160} & \textbf{0.191} & \textbf{0.976}\\ \hline
\multirow{2}{*}{Autoformer}  & ERM & 0.398 & 0.483 & 0.223\\
                             & \solution & \textbf{0.221} & \textbf{0.257} & \textbf{0.921}\\ \hline\hline
\end{tabular}
\egroup
\label{tbl:case}
\end{table}


\section{Conclusion and Discussion}
\label{sec:conclusion}

In this paper, we introduce a novel time-series problem \problem that addresses the challenge of achieving accurate forecasts in the presence of performativity. We present a natural solution to \problem, termed \solution, which effectively integrates the notion of delayed responses into time-series forecasting. By incorporating performativity, our proposed approach demonstrates notable improvements over conventional time-series forecasting models, as substantiated by both theoretical analyses and empirical investigations. Specifically, we have showcased the advantages of \solution through comprehensive experiments conducted on real-time COVID-19 and METR-LA traffic prediction tasks. Furthermore, our framework exhibits enhanced generalization capabilities and a heightened ability to capture underlying trends.

Beyond the context of \solution, it is crucial to recognize that delayed responses are not the only solution for anticipating distribution shifts in \problem. Recent research avenues~\cite{kulynych2022causal, miller2020strategic} explore alternative interpretations of performativity through the lens of causality. Thus, there exists the exciting prospect of investigating how causal discovery techniques, such as causal discovery, can assist in identifying the performativity in time-series forecasting problems. We expect that future works delve into these open questions, both theoretically and practically, to enrich our understanding of \problem.

\section{Acknowledgment}

This paper was supported in part by the NSF (Expeditions CCF-1918770, CAREER IIS-2028586, Medium IIS-1955883, Medium IIS-2106961, Medium IIS-2403240, PIPP CCF-2200269), CDC MInD program, Meta faculty gift, and funds/computing resources from Georgia Tech and GTRI.

\balance 
\bibliographystyle{ACM-Reference-Format}
\bibliography{sample-base}

\appendix

\section{Appendix I: \solution Analysis}
\label{sec:appenda}

To investigate the benefits of utilizing \solution\ in addressing \problem\ compared to conventional forecasting models through optimizing ERM, we analyze the discrepancy in the tightness of the generalization error bounds between the two approaches. In summary, our analysis presents two primary propositions: Firstly, when the $\bm{\mathrm{X}}^{DR}$ (typically refers to future data of features) is accessible, employing $\bm{\mathrm{X}}^{DR}$ for forecasting yields a tighter error bound compared to using $\pmb{x}$ (Proposition \ref{prop1}). Secondly, in scenarios where the $\bm{\mathrm{X}}^{DR}$ is unavailable and forecasting using the estimated $\hat{\bm{\mathrm{X}}}\vphantom{\bm{\mathrm{X}}}^{DR}$, there exists a trade-off in the error bound between reducing empirical loss and increasing model complexity as compared to forecasting with $\bm{\mathrm{X}}^{L}$ (Proposition \ref{prop2}). We now present a formal and detailed analysis of these propositions as follows.

For deriving the generalization error bounds for time-series data, the common practice assumes a $\beta$-mixing distribution for the time series $\bm{\mathcal{Z}}$ (e.g., $\bm{\mathcal{Z}}$ as the general representation of time series $\bm{\mathrm{X}}$ or $\bm{\mathrm{Y}}$). This property states that two distinct sequences become more independent as their time gap increases, which is naturally reasonable for real-world time-series data. When this assumption holds, the generalization bound for probably approximately correct (PAC) for time-series models follows:
\begin{theorem}
\label{thm:pac}
    ~\cite{mohri2008rademacher,kuznetsov2017generalization} Let $\Theta$ be the space of candidate predictors $\theta$ and let $\mathcal{H}$ be the space of induced losses: 
    \begin{gather*}
        \mathcal{H} = \{h=\mathcal{L} (\theta): \theta \in \Theta\}
    \end{gather*}
    for some loss function $0\leq \mathcal{L}(\cdot) \leq M$. Then for any sequential data $\boldsymbol{\mathcal{Z}}$ of size $n$ drawn from a stationary $\beta$-mixing distribution, with probability at least $1-\eta$:
    \begin{gather*}
        \mathbb{E} [\mathcal{L}(\theta)] \leq \mathcal{L}(\theta) + \delta(\mathcal{R}(\mathcal{H}), n, \eta)
    \end{gather*}
    where $\mathcal{R}(\cdot)$ measures the complexity of the model class $\Theta$, such as Rademacher Complexity, $\delta(\cdot)$ is a function of the complexity $\mathcal{R}(\cdot)$, the confidence level $\eta$, and the number of observed data points $n$.
\end{theorem}


\par\noindent\textbf{Proof of Theorem \ref{thm:pac}.} We first give the formal definition of stationary, $\beta$-mixing, and Rademacher complexity.

\par\noindent\textbf{Definition} (Stationary). A sequence of random variables $\boldsymbol{\mathcal{Z}} = \{\mathcal{Z}_t\}_{t=-\infty}^\infty$ is is said to be stationary if for any $t$ and and non-negative integers $m,k$, the random vectors $(\mathcal{Z}_t, \ldots, \mathcal{Z}_{t+m})$ and $(\mathcal{Z}_{t+k}, \ldots, \mathcal{Z}_{t+k+m})$ have the same distribution.

\par\noindent\textbf{Definition} ($\beta$-mixing). Let $\boldsymbol{\mathcal{Z}}=\{\mathcal{Z}_t\}_{t=-\infty}^\infty$ be a sequence of random variables. For any $i,j \in \mathbb{Z}$, let $\sigma_i^j$ denote the $\sigma$-algebra generated by the random variables $\mathcal{Z}_k$, $i\leq k \leq j$. Then for any positive integer $k$, the $\beta$-mixing coefficient of the stochastic process $\boldsymbol{\mathcal{Z}}$ is defined as
\begin{gather}
    \beta(k) = \sup_n \mathbb{E}_{B\in \sigma_{-\infty}^n}[\sup_{A\in \sigma_{n+k}^\infty}|P(A|B)-P(A)|]
\end{gather}
Then $\boldsymbol{\mathcal{Z}}$ is said to be $\beta$-mixing if $\beta(k) \rightarrow 0$ as $k\rightarrow \infty$.

\par\noindent\textbf{Definition} (Rademacher Complexity). Let $\boldsymbol{\mathcal{Z}}=\{\mathcal{Z}_t\}_{t=1}^n$ be the samples drawn from the distribution $\Pi$, then the empirical Rademacher complexity is
\begin{gather}
    \hat{\mathcal{R}}(\mathcal{H}) = \frac{2}{n} \mathbb{E} [\sup_{h\in\mathcal{H}} \bigg| \sum_{i=1}^n\sigma_i h(\mathcal{Z}_i)\bigg||\boldsymbol{\mathcal{Z}}]
\end{gather}
where $\sigma_i$ is a sequence of Rademacher random variables, independent of each other and everything else, and drawn randomly from $\{\pm 1\}$. $\mathcal{H}$ be the space of induced losses contains all possible $h$. The Rademacher complexity is 
\begin{gather}
    \mathcal{R}_n(\mathcal{H}) = \mathbb{E}_\Pi [\hat{\mathcal{R}}_n(\mathcal{H})]
\end{gather}

We first start with the stationary assumption holds, in which we assume the time series are both stationary and $\beta$-mixing. Then the PAC bound holds by the following lemma:

\begin{lemma}~\cite{mohri2008rademacher}
\label{thm:stat}
    Let $\Theta$ be the space of candidate predictors $\theta$ and let $\mathcal{H}$ be the space of induced losses: 
    \begin{gather*}
        \mathcal{H} = \{h=\mathcal{L} (\theta): \theta \in \Theta\}
    \end{gather*}
    for some loss function $0\leq \mathcal{L}(\cdot) \leq M$. Then for any sequential data $\boldsymbol{\mathcal{Z}}$ of size $n$ drawn from a stationary $\beta$-mixing distribution, and for any $\mu, k >0$, with $2\mu k = n$, and $\eta > 2(\mu -1 )\beta(k)$, where $\beta(k)$ is the mixing coefficient, with probability at least $1-\eta$:
    \begin{gather*}
        \mathbb{E} [\mathcal{L}(\theta)] \leq \mathcal{L}(\theta) + \hat{\mathcal{R}}_\mu(\mathcal{H}) + 3M \sqrt{\frac{\ln 4 /\eta^\prime}{2\mu}}
    \end{gather*}
    where $\eta^\prime = \eta - 4(\mu-1)\beta(k)$ and $\hat{\mathcal{R}}(\cdot)$ is the empirical Rademacher complexity.
\end{lemma}

We then remove the stationary assumption, in which we only assume the time series are $\beta$-mixing. Then the PAC bound is loosened by the following lemma:

\begin{lemma}~\cite{kuznetsov2017generalization}
\label{thm:nonstat}
    Let $\Theta$ be the space of candidate predictors $\theta$ and let $\mathcal{H}$ be the space of induced losses: 
    \begin{gather*}
        \mathcal{H} = \{h=\mathcal{L} (\theta): \theta \in \Theta\}
    \end{gather*}
    for some loss function $0\leq \mathcal{L}(\cdot) \leq M$. Then for any sequential data $\boldsymbol{\mathcal{Z}}$ of size $n$, drawn from a $\beta$-mixing distribution, and for any $\mu, k >0$, with $2\mu k = n$, and $\eta > 2(\mu -1 )\beta(k)$, where $\beta(k)$ is the mixing coefficient, with probability at least $1-\eta$:
    \begin{align*}
        \mathbb{E} [\mathcal{L}(\theta)] &\leq \mathcal{L}(\theta) + \frac{2}{k}\hat{\mathcal{R}}_\mu(\mathcal{H}) + \frac{2}{n} \sum_{t=1}^n \Bar{d}(t,n+s) \\
        &+ M \sqrt{\frac{\ln 2 /\eta^\prime}{8\mu}}
    \end{align*}
    where $\hat{d}$ is the averaged discrepancy, $s$ is the forecasting steps, $\eta^\prime = \eta - 4(\mu-1)\beta(k)$ and $\hat{\mathcal{R}}(\cdot)$ is the empirical Rademacher complexity.
\end{lemma}

Lemma \ref{thm:nonstat} provides a more general PAC bound of time-series data without the stationary assumption necessarily holding.  Lemma \ref{thm:nonstat} reduces exactly to Lemma \ref{thm:stat} if the stationary assumption holds, where the expected discrepancy is supposed to be 0 for stationary time series. 

By concluding Lemma \ref{thm:stat} and Lemma \ref{thm:nonstat}, the general form of the PAC bound for time series is given by:
\begin{gather*}
    \mathbb{E} [\mathcal{L}(\theta)] \leq \mathcal{L}(\theta) + \delta(\mathcal{R}(\mathcal{H}), n, \eta)
\end{gather*}
Then Theorem \ref{thm:pac} is proved.

In a more intuitive sense, $\delta(\cdot)$ refers to the non-vanishing generalization errors caused by the model complexity. For models with unknown complexity, such as autoregressive moving average (ARMA), the stationary assumption is necessary for leading to Lemma \ref{thm:stat}. However, for models with known complexity, such as neural networks, which are leveraged by \solution, the stationary assumption is not necessarily needed for the generalization bounds. Bridging Theorem \ref{thm:pac} to \solution, it is challenging to precisely quantify the exact model complexity for neural networks. However, analyzing disparities in empirical loss is still possible. In \solution, the delayed responses that maximize the similarity to the target are leveraged as forecasting inputs, which results in the following theorem:
\begin{theorem}
\label{thm:reg}
    Let $i\in[0,H]$ be all possible time delay and $\Theta=\{\theta_i\}$ be the time-series predictors for the corresponding $i$, including $\tau$. Let $\mathcal{L}$ be MSE-loss and $\mathcal{H} = \{h = \mathcal{L}(\theta_i): \theta_i \in \Theta\}$ be the space of induced losses. Then $\mathcal{L}(\theta_\tau) \leq \mathcal{L}(\theta_i), \forall i \in [0,H]$.
\end{theorem}
Here, $\theta_i$ reduces to conventional forecasting models when $\tau=0$. Intuitively, Theorem \ref{thm:reg} signifies that utilizing sequences most similar to the targets as forecasting inputs yields the lowest empirical error. 

\par\noindent\textbf{Proof of Theorem \ref{thm:reg}} 
Let $\bm{\mathrm{X}}$ denote the input features and $\bm{\mathrm{Y}}$ denote the target. Let $g$ be the linear regressor such that $g: \bm{\mathrm{X}}\rightarrow\bm{\mathrm{Y}}$. Then the optimal solution 
\begin{equation*}
    g^* = (\bm{\mathrm{X}}^T\bm{\mathrm{X}})^{-1} \bm{\mathrm{X}}^T \bm{\mathrm{Y}} = \frac{\bm{\mathrm{X}}^T\bm{\mathrm{Y}}}{\|\bm{\mathrm{X}}\|^2}
\end{equation*}

Then the empirical mean-square-error is $\|\bm{\mathrm{Y}} - \bm{\hat{\mathrm{Y}}}\|^2 = \|\bm{\mathrm{Y}} - \frac{\bm{\mathrm{X}}^T\bm{\mathrm{Y}}\bm{\mathrm{X}}}{\|\bm{\mathrm{X}}\|^2}\|^2$, denoted as $h(\bm{\mathrm{X}})$. 
According to the fact that $<\bm{\mathrm{X}}, \bm{\mathrm{Y}}> = \|\bm{\mathrm{X}}\|\|\bm{\mathrm{Y}}\|\cos\theta$,
then we can simplify $h(\bm{\mathrm{X}})$ as
\begin{equation*}
    h(\bm{\mathrm{X}}) = \|\bm{\mathrm{Y}}\|^2 (1-\cos^2\theta) = \|\bm{\mathrm{Y}}\|^2 \sin^2 \theta := h(\theta)
\end{equation*}
Then we have the derivate of $h(\theta)$ w.r.t. $\theta$ as
\begin{equation*}
    \frac{dh(\theta)}{d\theta} = =\|\bm{\mathrm{Y}}\|^2 \sin 2\theta
\end{equation*}
Suppose $\theta \in [-\pi, \pi]$, then it is evident that $h(\theta)$ has the local minimums when $\theta=0,\pm \pi$, that is, when the similarity is maximized. Hence, we can conclude that the MSE loss will reduce as the absolute similarity between the two signals increases, then Theorem \ref{thm:reg} is proved.

Combining Theorem \ref{thm:pac} and Theorem \ref{thm:reg}, we can conclude Proposition~\ref{prop1} and Proposition~\ref{prop2}.

\section{Appendix II: Evaluation Setup and Additional Results (Cont.)}
\label{sec:appendb}

\par\noindent\textbf{Additional Experiment Setups.}
In our evaluation of both regular time-series forecasting and real-time forecasting tasks using the COVID-19 dataset, we employ ensemble results derived from four distinct random seeds. This is because of the large variability observed in COVID-19 predictions through different seeds, and utilizing ensemble results helps reduce the uncertainty. Conversely, for METR-LA traffic predictions, we rely on predictions generated from a fixed random seed, as these predictions tend to exhibit stability across different random seeds.

In the case of the COVID-19 dataset, evaluation results are exclusively based on data from California (CA), Texas (TX), New York (NY), Florida (FL), and national data. Despite the model being trained on samples from all U.S. states, this selection is made for statistical significance considerations. For instance, evaluations of other states, like Alaska, are less statistically significant due to their smaller population bases or irregular reporting frequencies, such as biweekly data updates.

In the context of regular time-series forecasting tasks, all experiments follow a two-stage process: initially, we fine-tune all hyperparameters on the validation set, where all models are trained based on the training set. We retain all tuned hyperparameters fixed and subsequently retrain the model using joint training and validation sets. The testing results are then recorded based on the retrained models. We use such training schemes to mitigate the difference of time delays $\tau$ on training, validation, and test sets.

\begin{table}[H]
\centering
\caption{Performance comparison between forecasting using original inputs and exact $\bm{\mathrm{X}}^{DR}$ (marked as \ding{56} and \ding{52} respectively). Forecasting using $\bm{\mathrm{X}}^{DR}$ significantly improves the forecasting accuracy and correlations except with Autoformer.}
\bgroup
\def\arraystretch{1.05}
\resizebox{0.43\textwidth}{!}{%
\begin{tabular}{cc|ccc}
\hline\hline
\multicolumn{2}{c}{}         & \multicolumn{3}{c}{\texttt{\textbf{traffic}}}\\
Model                        & $\bm{\mathrm{X}}^{DR}$ & NMAE($\Downarrow$) & NRMSE($\Downarrow$) & PC($\Uparrow$) \\ \hline
\multirow{2}{*}{RNN}         & \ding{56} & 0.141 & 0.208 & 0.681 \\
                             & \ding{52} & \textbf{0.059} & \textbf{0.088} & \textbf{0.880}\\ \hline
\multirow{2}{*}{LSTNet}      & \ding{56} & 0.135 & 0.196 & 0.663 \\
                             & \ding{52} & \textbf{0.060} & \textbf{0.087} & \textbf{0.895}\\ \hline
\multirow{2}{*}{Transformer} & \ding{56} & 0.145 & 0.216 & 0.648\\
                             & \ding{52} & \textbf{0.062} & \textbf{0.091} & \textbf{0.898}\\ \hline
\multirow{2}{*}{Informer}    & \ding{56} & 0.141 & 0.198 & 0.673 \\
                             & \ding{52} & \textbf{0.062} & \textbf{0.093} & \textbf{0.870}\\ \hline
\multirow{2}{*}{Autoformer}  & \ding{56} & \textbf{0.166} & \textbf{0.217} & \textbf{0.530}\\
                             & \ding{52} & 0.181 & 0.239 & 0.450\\ \hline\hline
\end{tabular}}
\egroup
\label{tbl:ideal}
\end{table}

\section{Appendix III: Additional Evaluation Results}

\par\noindent\textbf{Forecasting with $\bm{\mathrm{X}}^{DR}$ yeilds tighter error bound.} 
We here show the additional results to validate Proposition~\ref{prop1}, which compares the performance disparities between forecasting with original inputs and with exact $\bm{\mathrm{X}}^{DR}$. The results of evaluations on METR-LA traffic dataset are shown in Table~\ref{tbl:ideal}. Here, we exclude the evaluations on the COVID-19 dataset, as the COVID-19 dataset includes both performative and non-performative features which may perturb testing the impacts of forecasting with $\bm{\mathrm{X}}^{DR}$ from performative features.


The validation results show clear benefits of forecasting with the delayed response $\bm{\mathrm{X}}^{DR}$ than the original inputs. Particularly, we notice an averaged 56.8\% NMAE reduction, 56.0\% NRMSE reduction, and 32.9\% PC increase when forecasting with RNN, LSTNet, Transformer, and Informer. However, when employing Autoformer as the forecasting model, we observe a slight performance degradation. This is because of the mechanism of Autoformer which enforces the forecasting trend based on historical look-back windows for long-term stability, which limits its ability to learn rapid trend changes in the short term and aligns less with our focus on short-term forecasting tasks.


\par
\begin{figure}[H]
    \centering
    \includegraphics[width=0.95\linewidth]{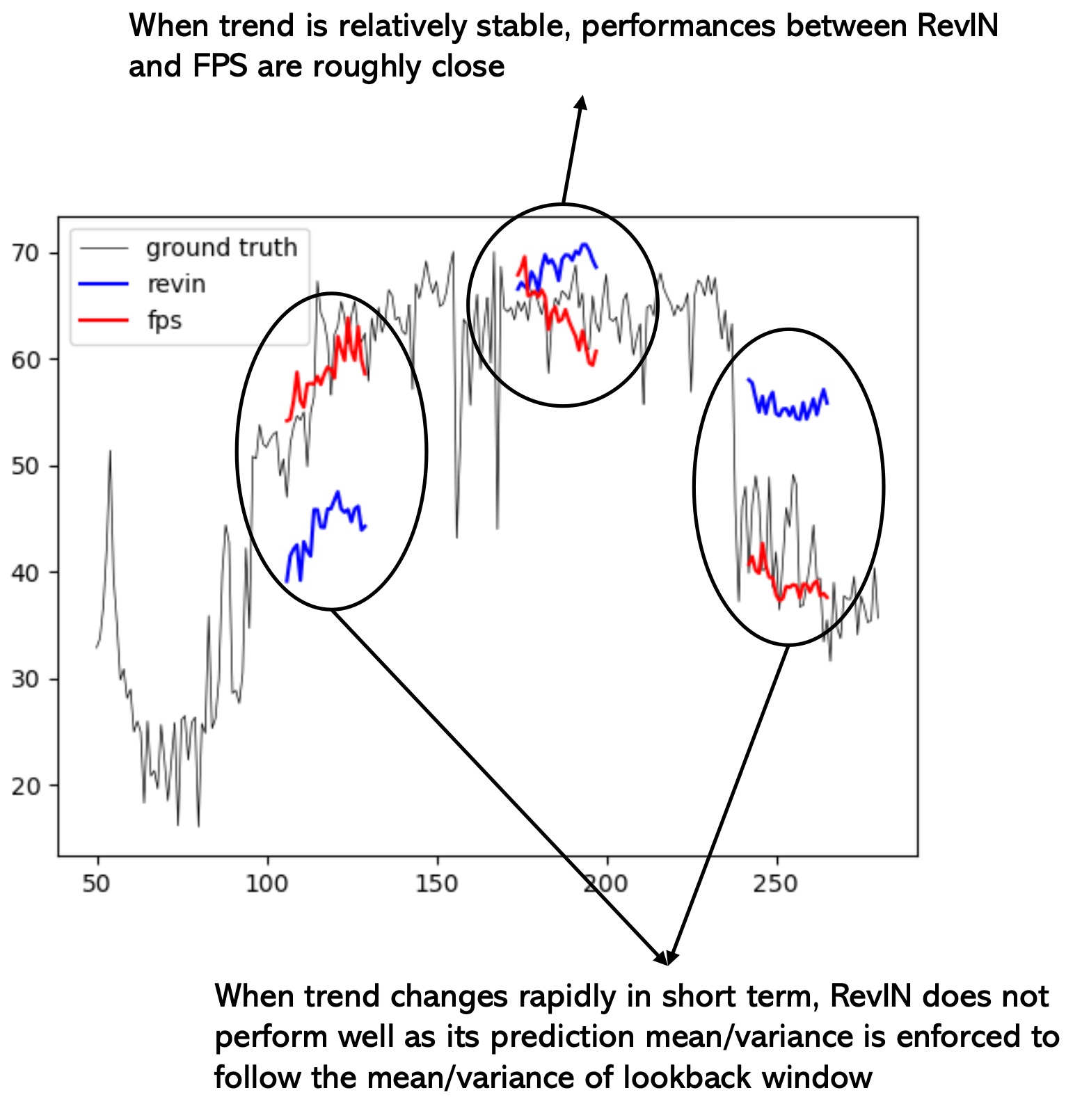}
    \caption{Visualizations of predictions from RevIN and \solution show that the normalization and de-normalization strategy in RevIN restricts the model's ability to capture the correct trend changes in the short term.}
    \label{fig:revin}
\end{figure}

\noindent\textbf{Non-stationarity methods can be detrimental to short-term forecasting.} As shown in Table~\ref{tbl:ts}, forecasting results using non-stationary methods, such as RevIN and Dish-TS, often do not lead to performance improvements and may even degrade performance in most scenarios. The intuition behind this is that existing non-stationary methods focus on producing stable predictions, which contrasts with the design of \solution, which aims to capture rapid trend changes caused by performativity. Non-stationary methods typically employ normalization and de-normalization to ensure that predictions within the forecasting window follow a similar mean and variance as the lookback window. While these approaches help maintain stable forecasting results in the long term, they restrict the exploration of trend information and enforce stability in predictions. This interpretation is further supported by visualizations of predictions on the METR-LA dataset in Figure~\ref{fig:revin}.

\end{document}